\PassOptionsToPackage{breaklinks}{hyperref}
\PassOptionsToPackage{dvipsnames, x11names}{xcolor}
\documentclass[sigconf]{acmart}

\newcommand{\paperTitle}{Evaluating Generalizability of Fine-Tuned Models for Fake News Detection}
\newcommand{\paperKeywords}{Generalizability, fine-tuning, fake news, covid-19, concept drift}
\newcommand{\paperAuthors}{Abhijit Suprem, Calton Pu}

\usepackage{endnotes,microtype,xspace,graphicx,fancyvrb,multirow}
\usepackage{pifont}
\usepackage{amsmath,amsopn}
\usepackage{listings}
\usepackage{subcaption}
\usepackage{mathrsfs}
\usepackage{booktabs}
\usepackage{listings}
\usepackage[labelfont=bf,font=small,textfont=md,belowskip=-5pt]{caption}
\usepackage[normalem]{ulem}
\useunder{\uline}{\ul}{}

\usepackage[breaklinks]{hyperref}
\hypersetup{%
    pdfauthor = {\paperAuthors},
    pdftitle = {\paperTitle},
    pdfkeywords = {\paperKeywords},
    bookmarksopen = {true},
    colorlinks=true,
    citecolor={urlcolor},
    linkcolor={linkcolor},
    urlcolor={citecolor},
    pdfborder={ 0 0 0 }
}

\usepackage[capitalize,noabbrev,nameinlink]{cleveref}

\newcommand{\squishitemize}{
\begin{list}{$\bullet$}
	{ \setlength{\itemsep}{0pt}
		\setlength{\parsep}{3pt}
		\setlength{\topsep}{3pt}
		\setlength{\partopsep}{0pt}
		\setlength{\leftmargin}{1.95em}
		\setlength{\labelwidth}{1.5em}
		\setlength{\labelsep}{0.5em} } }

\newcounter{Lcount}
\newcommand{\squishlist}{
	\begin{list}{\arabic{Lcount}. }
		{ \usecounter{Lcount}
			\setlength{\itemsep}{0pt}
			\setlength{\parsep}{3pt}
			\setlength{\topsep}{3pt}
			\setlength{\partopsep}{0pt}
			\setlength{\leftmargin}{2em}
			\setlength{\labelwidth}{1.5em}
			\setlength{\labelsep}{0.5em} } }
	
\newcommand{\squishend}{\end{list}}

\usepackage{xstring}
\newcommand{\PP}[1]{
\vspace{2px}
\noindent{\bf \IfEndWith{#1}{.}{#1}{#1.}}
}

\def\equationautorefname~#1\null{Eq. ~(#1)\null}

\def\Snospace~{\S{}}

\newcommand{\zerodisplayskips}{%
  \setlength{\abovedisplayskip}{0pt}%
  \setlength{\belowdisplayskip}{0pt}%
  \setlength{\abovedisplayshortskip}{0pt}%
  \setlength{\belowdisplayshortskip}{0pt}}
\appto{\normalsize}{\zerodisplayskips}
\appto{\small}{\zerodisplayskips}
\appto{\footnotesize}{\zerodisplayskips}

\AtBeginDocument{%
  \providecommand\BibTeX{{%
    \normalfont B\kern-0.5em{\scshape i\kern-0.25em b}\kern-0.8em\TeX}}}

\begin{document}

\title{\paperTitle}

\author{Abhijit Suprem}
\email{asuprem@gatech.edu}
\affiliation{%
  \institution{Georgia Institute of Technology}
  %\streetaddress{P.O. Box 1212}
  \city{Atlanta}
  \state{GA}
  \country{USA}
}

\author{Calton Pu}
\email{calton@cc.gatech.edu}
\affiliation{%
  \institution{Georgia Institute of Technology}
  %\streetaddress{P.O. Box 1212}
  \city{Atlanta}
  \state{GA}
  \country{USA}
}

\renewcommand{\shortauthors}{Suprem, et al.}

\begin{abstract}    
  The Covid-19 pandemic has caused a dramatic and parallel 
  rise in dangerous misinformation, denoted an `infodemic' 
  by the CDC and WHO. 
  Misinformation tied to the Covid-19 infodemic 
  changes continuously; this can lead to performance 
  degradation of fine-tuned models due to concept drift.
  Degredation can be mitigated if models generalize well-enough
  to capture some cyclical aspects of drifted data. 
  In this paper, we explore generalizability of pre-trained 
  and fine-tuned fake news detectors across 9 fake news datasets. 
  We show that existing models often overfit on their 
  training dataset and have poor performance on unseen data. 
  However, on some subsets of unseen data that overlap with training
  data, models have higher accuracy.
  Based on this observation, we also present KMeans-Proxy, 
  a fast and effective method based on K-Means clustering for 
  quickly identifying these overlapping subsets of unseen data.
  KMeans-Proxy improves generalizability on unseen fake news datasets by 
  0.1-0.2 f1-points across datasets. 
  We present both our generalizability experiments as well as 
  KMeans-Proxy to further research in tackling the fake news problem. 
  \end{abstract}

\maketitle

\section{Introduction}
\label{sec:intro}
The rapid spread of the Covid-19 virus has led
to a parallel surge in misinformation and disinformation \cite{misinfo}
This surge of false information, coined an `infodemic' by the 
CDC \cite{infodemic} can be life-threatening, 
destabilizing, and potentially dangerous 
\cite{infolife}. 
The infodemic is multimodal, meaning associated fake news 
can take forms of social media posts, tweets, 
articles, blogs, commentary, misrepresented titles and headlines, 
videos, and audio content.

\PP{Current Research.} 
There is significant progress on developing domain-specific 
automated misinformation detection and classification tools 
\cite{coaid, misinfo1, colies, fndetect, cosocial, generalization}. 
Such tools analyze labeled datasets in aforementioned modalities
to classify fake news. 
Recent approaches focus on transformer-based classifiers and language 
modelers \cite{generalization, fndetect}.

Such fake news detectors are specific to the datasets 
they are trained with \cite{colies, generalization}.
More recently, there is a focus on addressing generalizability 
concerns in these models 
\cite{generalization, generalization2, mcnnet, explainmisinfo, covidfn, fnft}. 
For example, \cite{generalization} explores the impact of generalization 
of 15 transformer models on 5 fake news datasets.
%
% [ref] studies shared task evaluation on seen and unseen 
% dataset pairs. 
%
The results show there is a generalizability gap in 
fake news detection: a fine-tuned model trained on 
one fake news dataset performs poorly on other 
unseen, but related, fake news datasets.

\PP{Generalization Study.}
In this paper, we study the generalizability and fine-tuning tradeoff 
and present our findings for furthering the research interest. 
We study several fake news detection architectures across
9 fake news text datasets of different modalities.
We find that fine-tuned models often have reduced accuracy 
on any unseen dataset.
However, when paired with a reject option to abstain from 
low-confidence predictions, fine-tuned models perform significantly better. 
These abstention results can then be labeled with 
active learning, crowdsourcing, weak label integration, 
or a variety of other methods present in literature. 

\PP{KMeans-Proxy}
Through observations on our generalizability results, we present a 
simple `reject option' \cite{reject1, reject2, reject3} for fake news
detectors, called KMeans-Proxy. 
KMeans-Proxy is based on KMeans clustering, and is inspired
by research into proxy losses \cite{proxy1, proxy2} and foundation models 
\cite{foundation1, foundation2, foundation3}.
It is written as a PyTorch layer and requires only a 
few lines of code to implement for most feature extractors. 
We show in our results that KMeans-Proxy improves 
generalization on fake news datasets by 0.1 to 0.2 
f1 points across several experiments.

\PP{Contributions} 
In summary, our contributions are:

\squishlist
\item	Extensive set of experiments across 9 fake news datasets on the generalizability/fine-tuning trade off.

\item KMeans-Proxy, a simple reject-option for feature extractors. KMeans-Proxy 
uses cluster proxies from ProxyNCA to estimate embedding cluster centers 
of the training data. 
During prediction, KMeans-Proxy provides a reject option based on 
label difference between sample prediction and nearest 
training data cluster center.
\squishend

Our code for running experiments and for KMeans-Proxy are provided\footnote{\url{https://github.com/asuprem/GLAMOR/blob/colabel-multicb/src/ednaml/utils/blocks/KMeansProxy.py}}.
\section{Related Work}
\label{sec:related}
%We first cover recent work in inherently interpretable models and \posthoc explanations. We will then cover vehicle feature extraction.

% \begin{figure}[t] 
%     \centering \includegraphics[width=3.4in]{figs/inherent}
%   \caption{\textbf{Constructive Interpretability}: Models
%   with constructive interpretability provide explanations for their predictions.
%   %
%   This is achieved by incorporating
%     human knowledge. Here, an interpretable model
%     explains its prediction is based on vehcle color, type, seats, and vehicle part similarities.
%    }
%   \label{fig:inherent}
%   \end{figure}

\subsection{Generalizability and Fine-Tuning}
\label{sec:generalizability}
Since 2002, there are increasing numbers of Covid-19 
fake news datasets and associated models for these datasets \cite{coaid, cosocial}.
Recently, there is increasing interest in gauging 
the effectiveness of each of these fine-tuned models 
on related, but unseen datasets \cite{generalization}. 
The authors of \cite{generalization} conduct a generalization study 
over 15 model architectures over 5 datasets, and find 
that fine-tuning offers little advantage in 
classification accuracy. 
%
% Similarly, [ref] studies generalization under paired 
% datasets, and finds domain mapping to be useful in 
% improving generalizability. 
%
There is also an abundance of research in unsupervised 
domain adaptation to recover accuracy under changing domains 
or concept drift \cite{odin, drift1, drift2, drift3}.

\subsection{Concept Drift}
Concept drift occurs when testing or prediction data 
exhibits distribution shift \cite{drift1}, either in the data domain, 
or in the label domain \cite{drift3}. 
Data domain shift can include introduction of new 
vocabularies, disappearance of existing words, 
and word polysemy \cite{litmus1}.
% , where a word can take new meanings 
% that can be markedly different from the original meaning [litmus]. 
%
Label domain shift occurs when the label space 
itself changes for the same type of data \cite{drift2, ebka}.
For example, when new types of misinformation are detected, 
then the boundary between misinformation and true 
information must be adjusted \cite{ebka}. 
%
% We show example of data domains shift in Fig [], 
% where fake news data exists in different clusters 
% that have some overlaps. [TODO for extension]
%
We show label shift in \cref{fig:overlap}, where subsets of true and 
fake news occupy the same embedding space across datasets 
due to fine-grained differences.

\subsection{Reject Options}
One drawback of classification models is that 
they provide a prediction for every data point \cite{reject1}, 
regardless of confidence. 
Reject options perform external or internal 
diagnosing.
This can help detect either low confidence due to 
low coverage or divergence from training data distributions 
due to concept drift. 
Several approaches are covered in a recent survey \cite{reject1}. 
We present a reject option that uses recent 
findings in \cite{foundation2} and \cite{niceness} on the topology 
of the embedding space:
(i) find that local smoothness of the 
label space is indicative of local accuracy and coverage \cite{niceness}, and 
(ii) local label shift, where 
nearby samples have different labels, is a good predictor 
of local smoothness \cite{foundation2}. 
Our reject option, described in \cref{sec:kmeans} is a clustering approach
that calculates cluster centers in the feature extractor embedding space
for the training data.
Then, during prediction, a model can provide a prediction as well 
the label for the nearest training data cluster center. 
Flipped, or different, labels can indicate reduced local smoothness, 
confidence, and coverage, leading to a reject decision.

\subsection{Motivation}
It is well known that fine-tuned models suffer performance 
degradation over time due to data domain shift \cite{ebka, argo, gft}. 
Usually, this performance degradation is detected, and a new 
model is trained on new labeled data. 
Recently, the velocity and size of new data makes obtaining 
labeled data quickly and at scale, very expensive \cite{mdaws}. 
Updating models during data domain shift requires relying 
on weak labels, authoritative sources, and hierarchical 
models \cite{mdaws, snorkel, eews}. 
In such cases, a team of prediction models is pruned and updated 
with new training data \cite{ebka}. 
However, we still need predictions in the period when data domain 
shift is occurring, and new models have not been trained.
Our work, as well as recent research in generalizability \cite{generalization}, 
weak labeling \cite{snorkel}, foundation models \cite{foundation2}, and rapid fake 
news detection \cite{mdaws} falls in this period. 
Our generalizability experiments in \cref{sec:genexp} show that fine-tuned 
models, while having lower performance on unseen data, 
do have better accuracy on some subsets. 
Our KMeans-Proxy solution finds these subsets where fine-tuned 
models have higher accuracy. 
\section{Generalization Experiments}
\label{sec:genexp}

We transformer-based text feature extractors for fake news 
classification generalizability with several experiments. 
We cover the datasets, architectures, and experiments below.

\PP{Datasets.}
We used 9 fake news datasets, consisting of blog articles, 
news headlines, news content, tweets, social media posts, 
and article headlines. 
We have described our dataset below in \cref{tab:datasets}. 

\begin{table}[t]
    \caption{Datasets used for our experiments. If splits were not available, we used a random class-balanced split. For Tweet datasets, sample counts are after rehydration, which removed some samples due to missing tweets.}
    \label{tab:datasets}
    \begin{tabular}{lrrr}
    \hline
    \textbf{Dataset} & \textbf{Training} & \textbf{Testing} & \textbf{Type}  \\ \hline
    k\_title \cite{kagglefn}          & 31k               & 9K               & Article Titles \\
    coaid \cite{coaid}                & 5K                & 1K               & News summary   \\
    c19\_text \cite{cov19fn}          & 2.5K              & 0.5K             & Articles       \\
    cq   \cite{covidcq}               & 12.5K             & 2K               & Tweets         \\
    miscov    \cite{miscov}           & 4K                & 0.6K             & Headlines      \\
    k\_text   \cite{kagglefn}         & 31k               & 9K               & Articles       \\
    rumor  \cite{covrumor}            & 4.5K              & 1K               & Social Posts   \\
    cov\_fn  \cite{covidfn}           & 4K                & 2K               & Tweets         \\
    c19\_title  \cite{cov19fn}        & 2.5K              & 0.5K             & Article Titles \\ \hline
    \end{tabular}
    \end{table}

\begin{figure}[t] 
    \centering \includegraphics[width=3.4in]{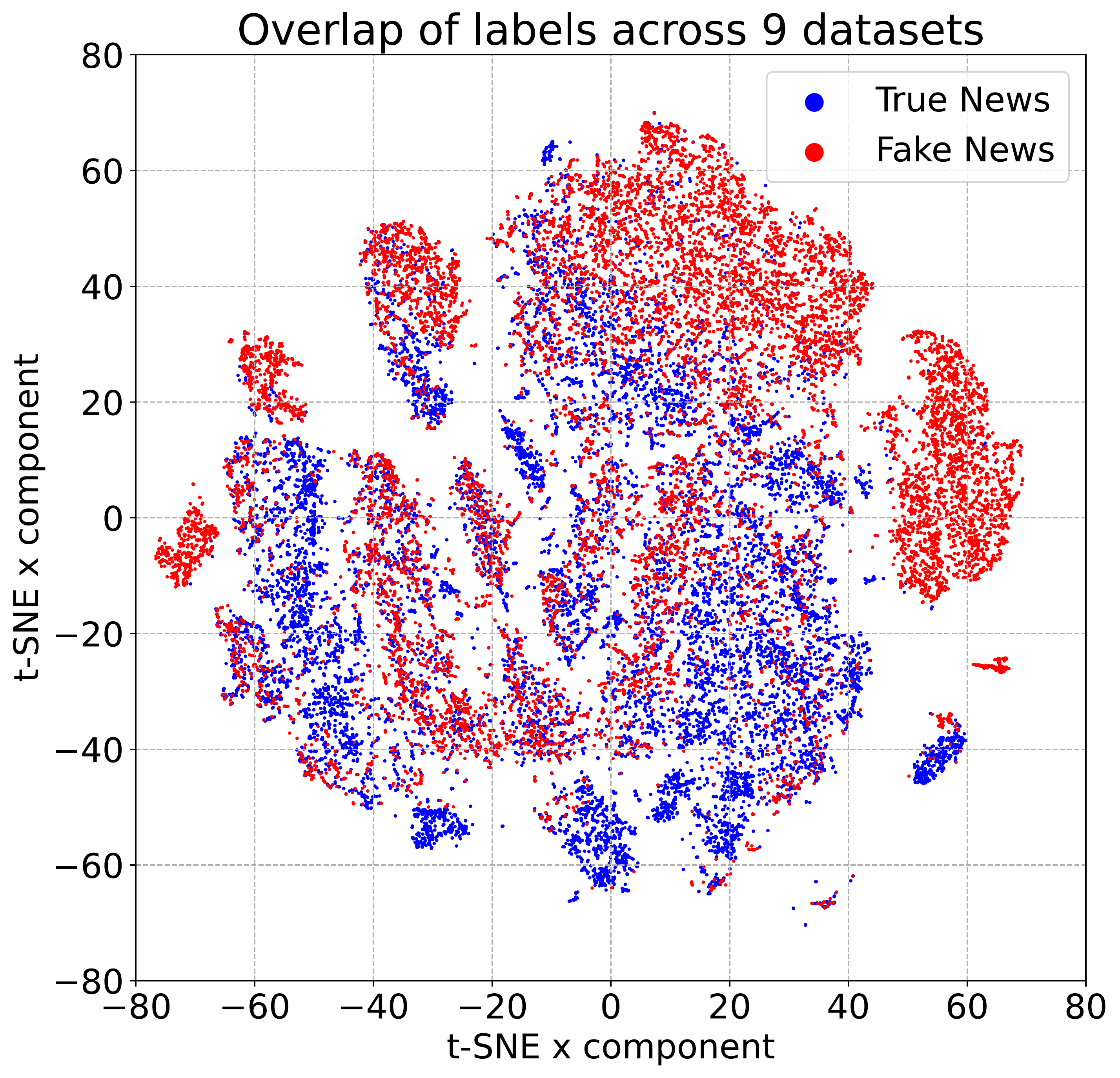}
  \caption{\textbf{Overlap of label embeddings}: We show label embeddings for all 9 datasets here. 
  There are several regions where true news (blue) overlap with fake news (red) across different datasets.
  This can make models for one dataset perform worse on unseen, but related text-based fake news datasets.}
  \label{fig:overlap}
  \end{figure}

Where possible, we have used the provided training and 
validation sets; otherwise, we performed a random, 
class-balanced 70-30 split for training and testing. 
For \cite{covidcq} and \cite{covidfn} datasets, we performed tweet 
rehydration, which removed some samples due to missing tweets. 
We show example of label shift due to label overlap in \cref{fig:overlap}.
Here, samples from each dataset are passed through a pre-trained
BERT classifier. 
The BERT embeddings are then reduced to 50 components with PCA
then to 2 components with tSNE.
There are several regions with label overlaps, where samples with positive 
and negative labels occupy similar spaces.

\PP{Architectures.}
We use BERT and AlBERT architectures for our experiments \cite{bert, albert}. 
Each transformer architecture converts input tokens to a classification feature vector.
We used pretrained architectures as starting points; our selections include
the BERT \cite{bert}, AlBERT \cite{albert}, and COVID-Twitter-BERT \cite{covbert}.

\PP{Experiments}
We performed 3 experiments to evaluate generalizability of covid fake 
news detectors. 
In each case, our starting point is a pre-trained foundation model, described in previous
section.
We then conduct the following experiments:

\squishlist
\item \textbf{Static-Backbone.} We freeze the pre-trained feature extractor backbone, 
and train only the classifier head. This is analogous to using a static foundation 
model.
\item \textbf{Static-Embedding.} We fine-tune the transformer part of the pre-trained 
feature extractor along with the classifier head together with a single optimizer, 
and freeze the embedding module
\item \textbf{Fine-Tuned Backbone.} We fine-tune the entire feature extractor backbone 
along with the classifier head.
\squishend

\PP{Evaluation.}
We average results across multiple runs of each transformer architecture. 
To show results in limited space, we have provided complete evaluation results
for backbones using Covid-Twitter-BERT.
To test generalizability, we train each model on a single dataset, and evaluate on the 
test-sets of the remaining, unseen datasets as well as its own testing dataset. 
Our results are presented as a confusion matrix. 
All approaches are trained for 5 epochs with an AdamW optimizer, with a 
learning rate of 1e-4, with a batch size of 64.

\subsection{Generalizability Results}
\begin{figure}[t] 
    \centering \includegraphics[width=3.4in]{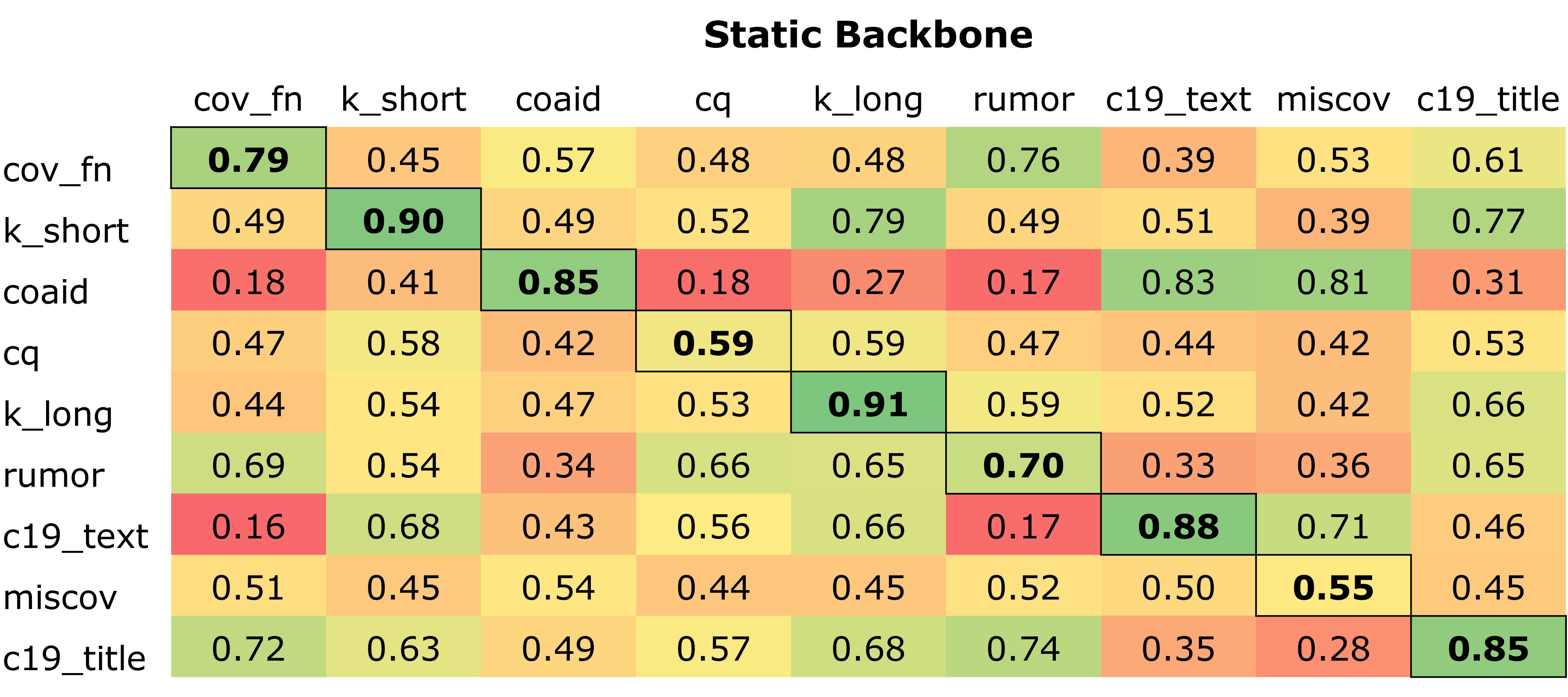}
  \caption{\textbf{Confusion Matrix for Static Backbone}}
  \label{fig:sbconf}
  \end{figure}

\begin{figure}[t] 
    \centering \includegraphics[width=3.4in]{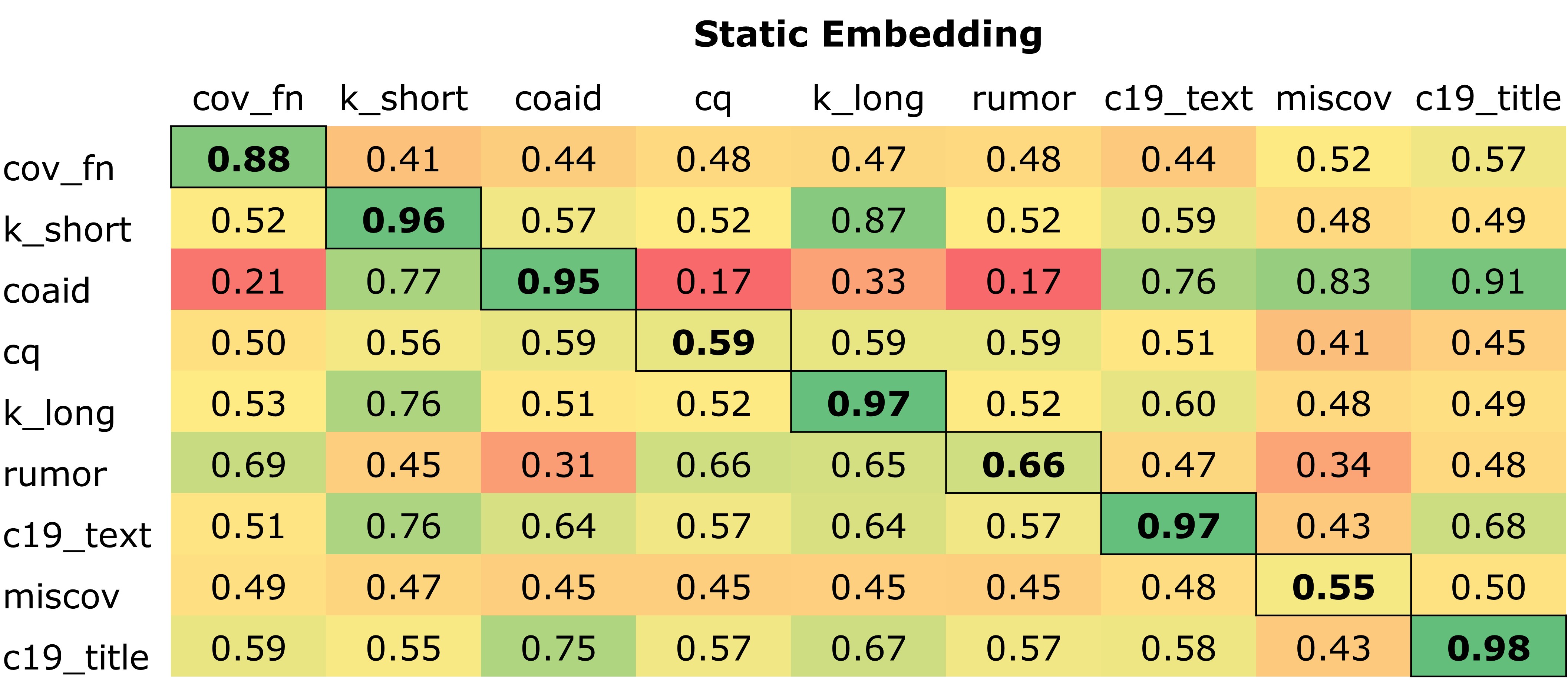}
  \caption{\textbf{Confusion Matrix for Static-Embedding Backbone}}
  \label{fig:seconf}
  \end{figure}

\begin{figure}[t] 
    \centering \includegraphics[width=3.4in]{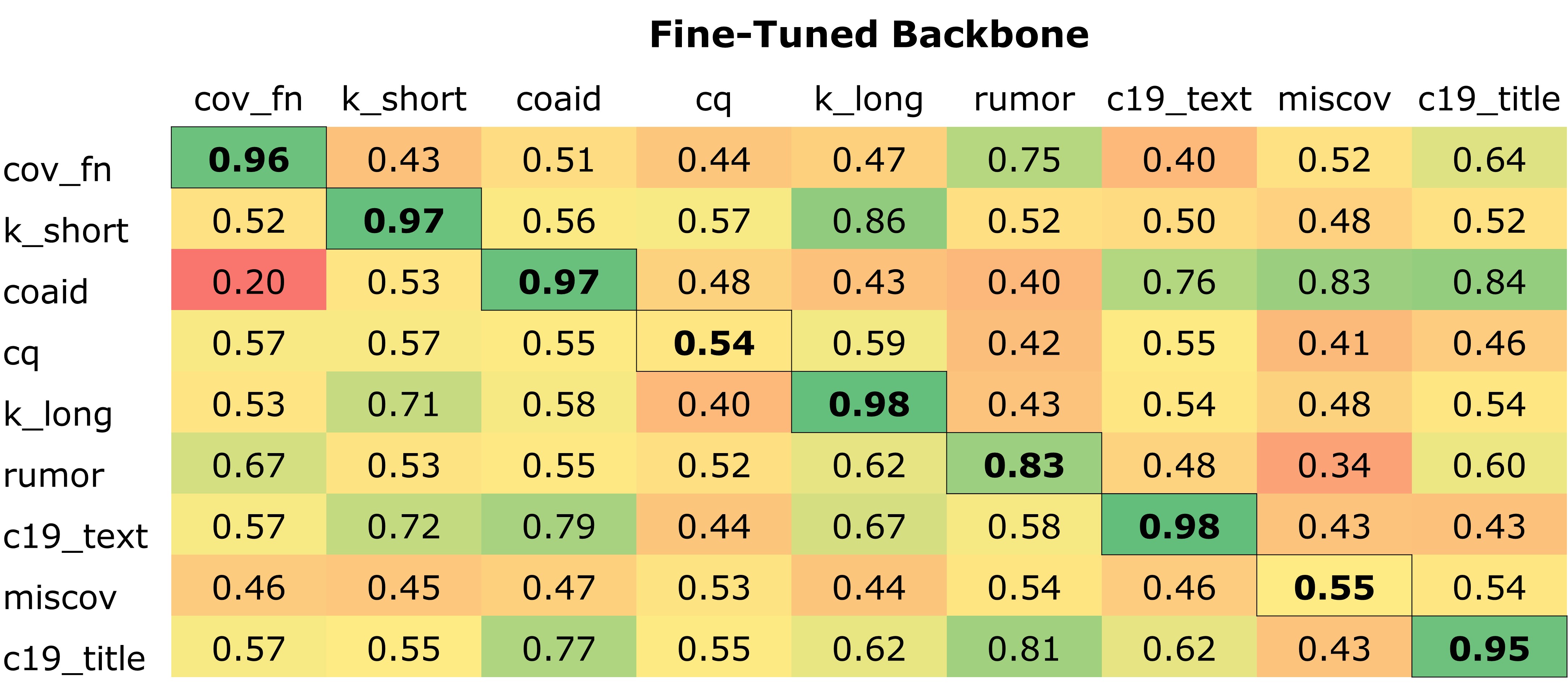}
  \caption{\textbf{Confusion Matrix for FIne-Tuned Backbone}}
  \label{fig:ftconf}
  \end{figure}

We show generalizability results using the COVID-Twitter backbone in 
for static-backbone training in \cref{fig:sbconf},
static-embedding training in \cref{fig:seconf},
and fine-tuned backbone training in \cref{fig:ftconf}.

\PP{Static backbone vs Fine-Tuning.}
The confusion matrices show that upon fine-tuning, 
each model increases accuracy on its corresponding test dataset. 
For example, accuracy on `k\_short' increases from 0.90 to 
0.97 between the static backbone and the fine-tuned backbone. 
However, this fine-tuning comes at the cost of generalization in some cases: 
across several datasets, model accuracy on unseen data suffers in 
the fine-tuned backbone experiments. 
For example, a model trained on `cov\_fn' achieves f1 of 0.72  when tested on `c19\_title' in the 
static-backbone experiment in \cref{fig:sbconf}. On the fine-tuned experiment,
the same trained model achieves f1 of 0.57, approximately a 20\% drop. 
SImilarly, a model trained on `miscov' and tested on `c19\_text' achieves f1 of 0.71
with static backbone, versus f1 of 0.43 with fine-tuned backbone.
This indicates once a model is fine-tuned on a specific covid dataset, 
it loses some generalization information compared to the static-backbone version.
However, this is not consistent.
In some cases, generalization accuracy increases: `rumor' performs
better on `cov\_fn' after fine-tuning, with f1 of 0.52 on the static backbone, 
versus an f1 of 0.75 after fine-tuning.
Furthermore, `rumor' achieves f1 of 0.67 when tested on `c19\_title` 
on the static backbone, and f1 of 0.81 on the fine-tuned backbone (conversely, it performs \textit{worse} on 
`coaid`, with f1 dropping from 0.83 to 0.40).
%
%Some datasets have overlap on their content, as we see from the tSNE in Figure []. 
%
% On these unseen datasets, the fine-tuned models perform better than their 
% static counterparts. 
%

\begin{figure*}[t]
  \centering
  \begin{subfigure}{0.24\textwidth}
      \includegraphics[width=\textwidth]{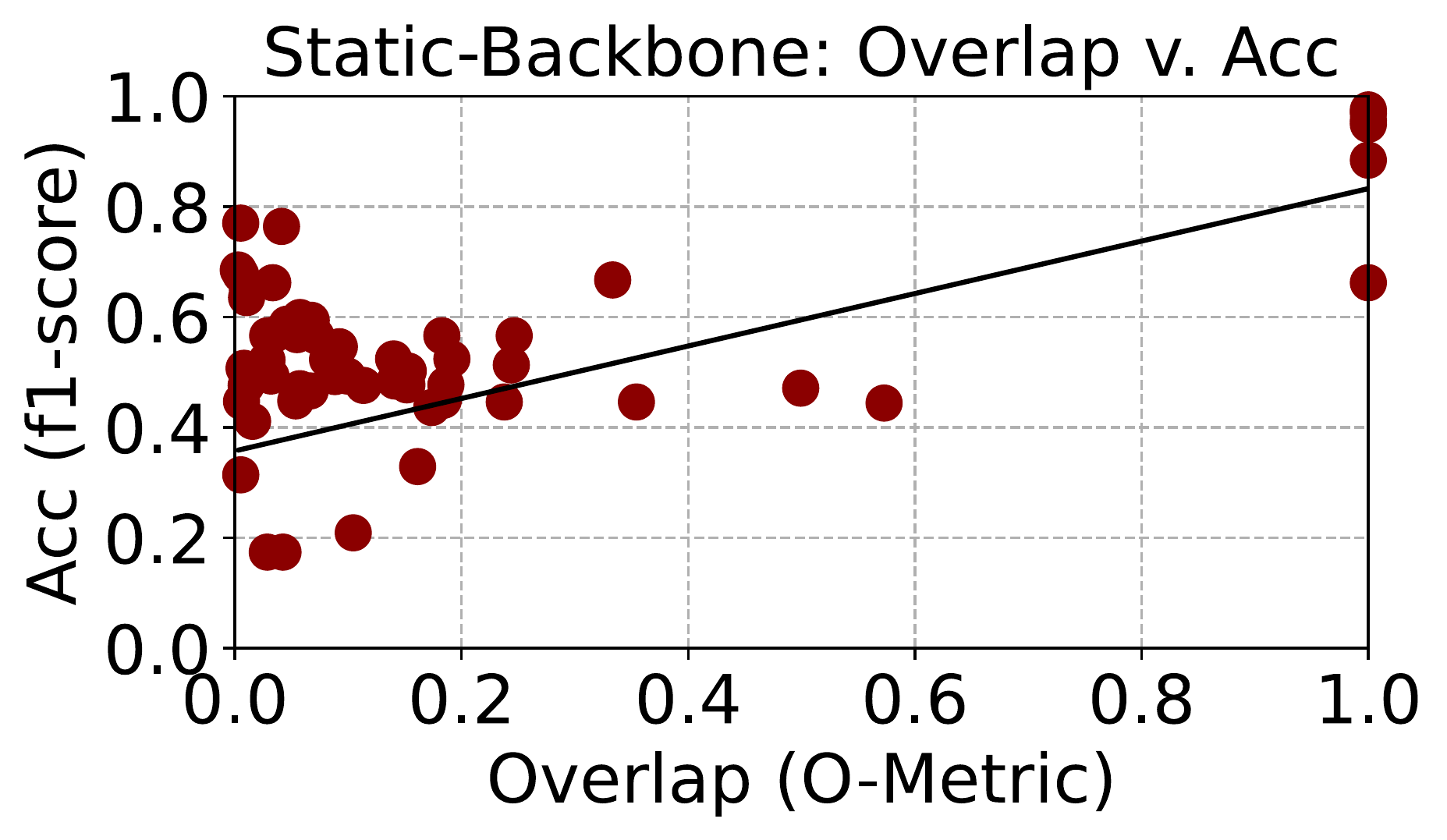}
      \caption{Static-Backbone}
      \label{fig:sb}
  \end{subfigure}
  \hfill
  \begin{subfigure}{0.24\textwidth}
      \includegraphics[width=\textwidth]{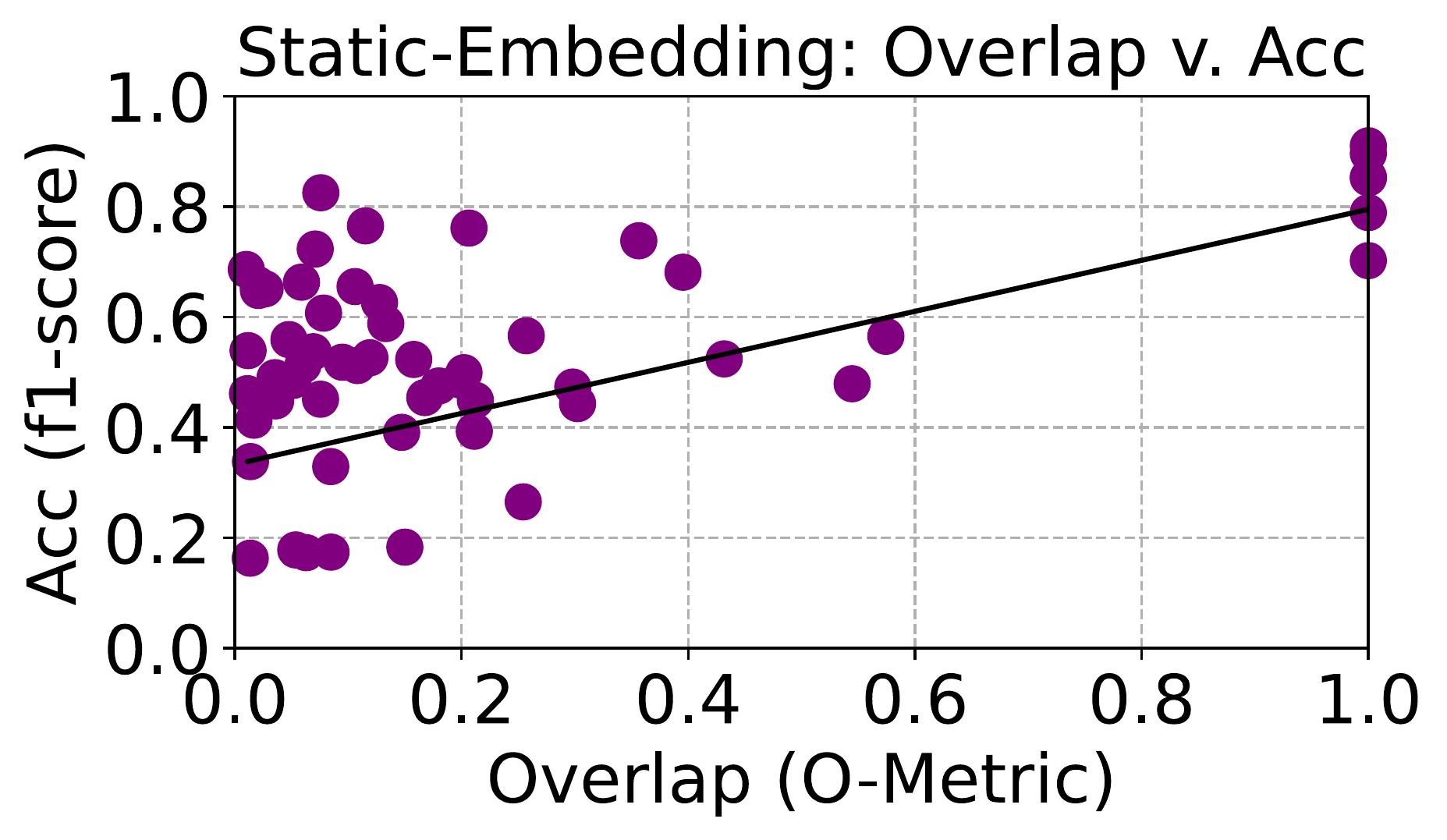}
      \caption{Static-Embedding}
      \label{fig:se}
  \end{subfigure}
  \hfill
  \begin{subfigure}{0.24\textwidth}
      \includegraphics[width=\textwidth]{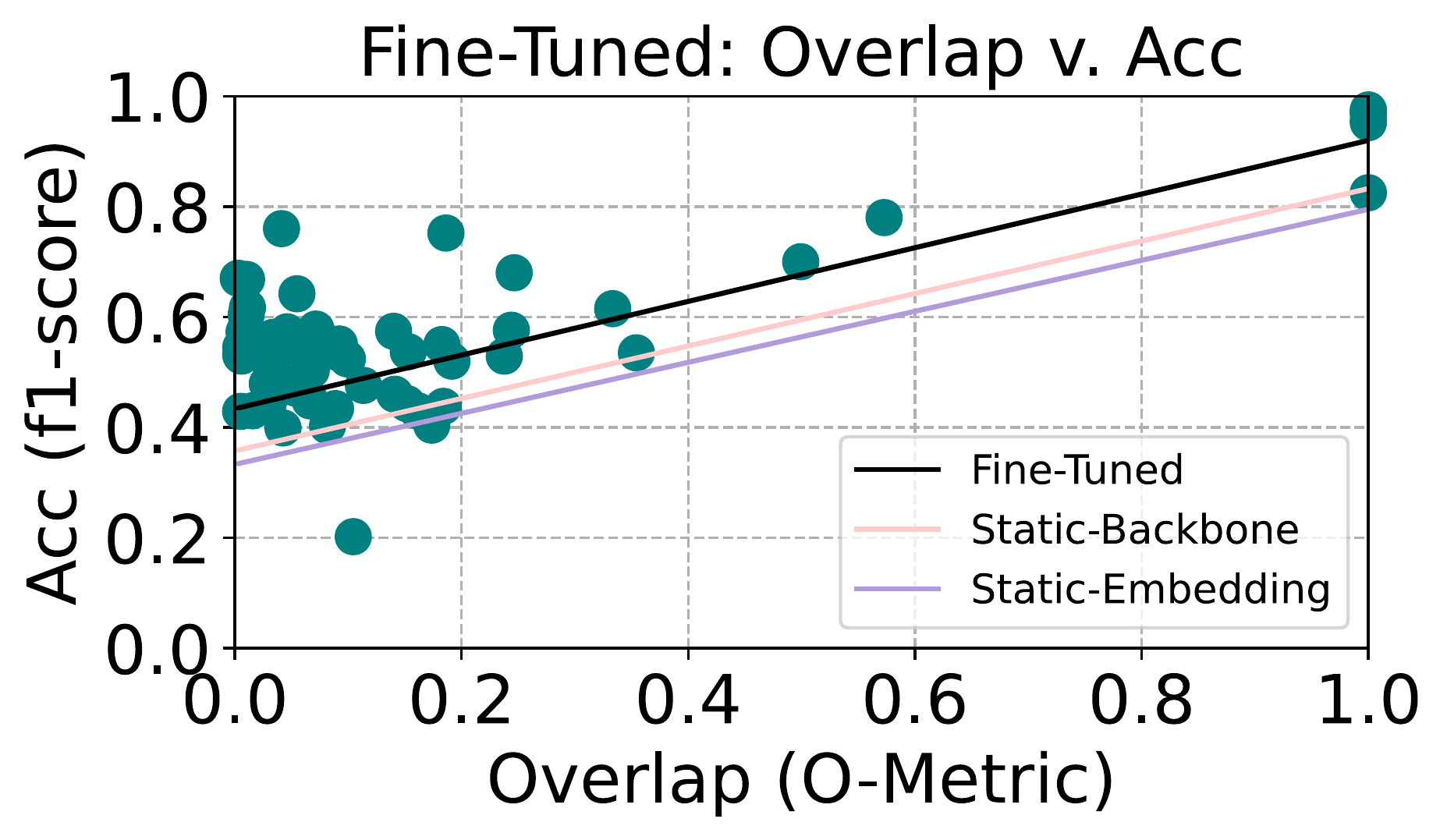}
      \caption{Fine-tuned}
      \label{fig:ft}
  \end{subfigure}
  \hfill
  \begin{subfigure}{0.24\textwidth}
      \includegraphics[width=\textwidth]{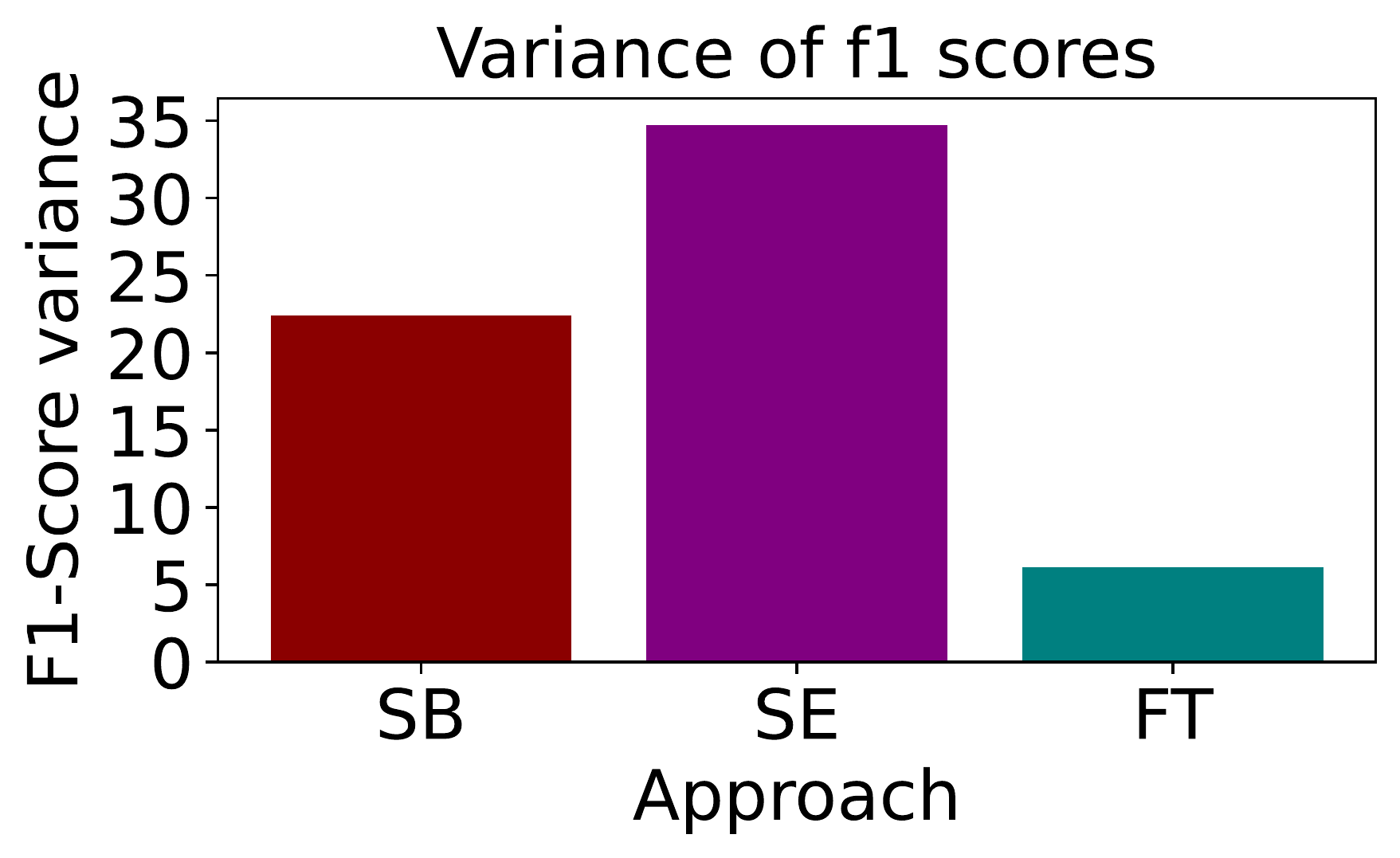}
      \caption{F1-Score Variance}
      \label{fig:variance}
  \end{subfigure}
          
  \caption{Overlap vs Accuracy for each of our experiment 
  approaches. In \cref{fig:ft}, we have also provided the 
  trendlines from the other experiments for comparison.}
  \label{fig:graphs}
  \end{figure*}

\PP{Static Embedding.}
A middle ground between complete fine-tuning and a fully static foundation 
feature extractor is to freeze the embedding layer and 
fine-tune the transformer layer of the backbone.
Recent work finds freezing the embedding layer during training
can reduce computation costs while achieving 90\% of the accuracy of a 
fully fine-tuned model 
\cite{sefreezing1, sefreezing2, sefreezing3, sefreezing4, seimportant}.
We find similar results, where freezing the embedding layer achieves accuracy
similar to the corresponding fine-tuned model on the testing dataset,
shown in \cref{fig:seconf}
However, on unseen data, accuracy drop has higher variance. 
It is not immediately clear what impact embedding freezing has on unseen
data accuracy.
For this, we must explore the actual overlap between datasets.

\subsection{Data Overlap and Accuracy}
We have seen that there is label overlap between
datasets in \cref{fig:overlap}.
We have also seen accuracy variances on unseen data:
rather than a linear drop across unseen data, some models
perform better and some perform worse after fine-tuning.
These can be explained by directly measuring dataset overlap.

\PP{O-Metric.}
To compute overlap, we use the O-metric to calculate
point-proximity overlap from \cite{ometric}.
The O-metric computes overlap between 2 sets of points in 
n-dimensional space using a distance-metric.
We find overlap as follows: given two datasets A and B, 
we compute the fraction of points in each dataset where
the nearest neighbor is not from the same dataset. 
So, for each point $x\in A$, first we obtain:

$$
O_A(x) = w_A(x) / b_A(x)
$$

where $w_A()$ is the distance to the nearest point to $x$ in A, 
and $b_A()$ is the distance to the nearest point to $x$ in B.
Then we can compute the ratio $p_A=|O_A>1|/|A|$ to find overlap
of B in A. 
$p_A$ is bounded in $[0,1]$; as $p_A$ approaches 1,
this indicates most points in $A$ are closer to a point in 
$B$ than in $A$.
The O-metric is bidirectional in computing overlap
 and includes both $p_A$ and $p_B$:

$O = \frac{p_A + p_B}{2}$

Since we are interested in evaluating generalization, where
we want to see only the overlap of unseen data on training data,
we use a directional O-metric.
That is, we let $O = p_A$ for the final overlap value
 in a context where A is the training dataset
and B is the unseen dataset.  
We compute the overlap value between each dataset pair using
cosine similarity on the embeddings of each data point.
So, for each model, we compute embeddings of every sample across
all 9 datasets, then compute the directional O-metric overlap
of each dataset on the model's training dataset [ALGO???].

\PP{Accuracy and Overlap}
We show in \cref{fig:graphs} the comparison of overlap to accuracy across
all 9 datasets on static-backbone, static-embedding, and fine-tuned backbone
experiments.
First, we see that as overlap increases, accuracy also increases
for the fine-tuned backbone in \cref{fig:ft}. 
Second, and perhaps more striking, there is higher variance in
accuracy on the static-embedding versus fine-tuned backbone models.
This fits the observations from \cite{seimportant} that the 
embedding layers capture more domain-specific information than
the deeper layers.
In this cases, because the embedding layers were frozen, they 
never learned the domain-specific fake news information.
Consequently, the static-embedding models' accuracy on unseen data 
sees significant variance, compared to the fine-tuned backbone 
and static-backbone models, shown in \cref{fig:variance}.

\PP{Generalization and Overlap.}
Clearly, higher overlap between evaluation data and training data 
is indicative of accuracy. 
During testing, however, it may be difficult to calculate this 
overlap on data for each sample.
Further, evaluation data changes continuously, so the overlap may 
itself change due to concept drift.
Recent research has shown the importance of exploring a model's 
feature space to identify embedding clusters \cite{foundation2, niceness}. 
These embedding clusters signify regions of the data space a 
model has captured. 
%
% Recent work in [ref] and [ref] have shown that a model's embedding space 
% is indicative of data domain coverage. 
% %
% A non-deterministic label region is one where positive and negative 
% labels overlap [survey]. 
% %
% This can occur in lower-dimensional projections, such as in our tSNE embeddings. 
% %
% A model that has good training data for such a region has higher 
% accuracy on these `flipped' labels [ref]. 
%
Metrics such as probabilistic Lipschitzness show that accuracy on 
embedding clusters 
can be bounded using the smoothness, or gradient, in the embedding space 
\cite{niceness}. 
Further, LIGER \cite{foundation2} shows that nondeterministic label regions,
i.e. where labels overlap, indicate non-smoothness.
We extend these findings to present KMeans-Proxy - a plug-and-play pytorch layer.

\subsection{KMeans-Proxy}
\label{sec:kmeans}
Intuitively, if we can store the coverage of a model's embedding space, 
then for any sample point, we can check if it falls inside the coverage. 
Further, we can also check if a model's prediction on the sample matches 
the prediction for the coverage. 
We can pair this with a coverage radius, e.g. by computing $r$ that constitutes
coverage of all points in a cluster that are 1 standard deviation away from the
cluster center with respect to a distance metric, such as the l2 norm.
Then, if the predictions do not match or a point falls outside the single standard deviation
coverage radius, this is a strong abstention/reject signal.

\PP{Implementation.} 
We can capture the coverage of the embedding space by 
using embedding proxies. 
Proxies are common in cluster and proxy NCA losses \cite{proxy1, proxy2}. 
We adapt them as the KMeans cluster centroids by acting as proxies for
the cluster centers.
This allows our approach to extend to online or continuous learning
domains as well.
The findings in \cite{foundation2} suggest increased partitioning of the embedding space 
can yield better local region coverage. 
So, KMeans-Proxy is initialized with 2 parameters: 
the number of classes $c$, and a proxy factor $k$. 
Then, we then obtain $k\cdot c$ centers, with $k$ proxies for each class, 
so that each cluster is a smaller, more representative local region.

\begin{lstlisting}
self.proxy = KMeansProxy(proxy_factor = 4, classes = 3, dimensions = 768)
\end{lstlisting}

During model training, KMeans-Proxy performs minibatch online KMeans clustering
 to obtain the embedding space proxies for cluster centers. Online clustering converges
 asymptotically, per \cite{kmeans}.
 
 \begin{lstlisting}
def forward(x):	# Proxy forward function
    if self.training:
        self.update_proxies(x)
        return x, None, None
    return x, self.nearest_proxy(x), self.nearest_proxy_label(x)  
\end{lstlisting}

During prediction, a model using KMeans-Proxy can predection,
as well as nearest proxy label and nearest proxy. 
A meta abstention policy can review for label flipping, or coverage radius.

\begin{lstlisting}
def forward(x):	 # Model forward function
    x = self.feature_extractor(x)
    x, proxy, proxy_label  = self.proxy(x)
    label = self.classifier(x)
    return label, proxy, proxy_label
\end{lstlisting}

% \usepackage{multirow}
% \usepackage[normalem]{ulem}
\begin{table}[t]
    \caption{Generalization improvement with KMP for model trained on `coaid'.}
    \label{tab:coaid}
    \begin{tabular}{|lrrrr|}
    \hline
    \multicolumn{5}{|c|}{\textbf{Trained on `coaid'}}                                                                                   \\ \hline
    \multicolumn{1}{|c|}{\multirow{2}{*}{\textbf{Testing Dataset}}} & \multicolumn{4}{c|}{\textbf{Approach}}                          \\ \cline{2-5} 
    \multicolumn{1}{|c|}{}                                          & SB            & SE            & FT            & \textbf{FT+KMP} \\ \hline
    \multicolumn{1}{|l|}{cov\_fn}                                   & \textbf{0.57} & 0.44          & 0.51          & {\ul 0.53}      \\
    \multicolumn{1}{|l|}{k\_short}                                  & 0.49          & {\ul 0.57}    & 0.56          & \textbf{0.57}   \\
    \multicolumn{1}{|l|}{coaid}                                     & 0.85          & 0.95          & {\ul 0.97}    & \textbf{0.98}   \\
    \multicolumn{1}{|l|}{cq}                                        & 0.42          & \textbf{0.59} & 0.55          & {\ul 0.57}      \\
    \multicolumn{1}{|l|}{k\_long}                                   & 0.47          & 0.51          & \textbf{0.58} & {\ul 0.55}      \\
    \multicolumn{1}{|l|}{rumor}                                     & 0.34          & 0.31          & {\ul 0.55}    & \textbf{0.68}   \\
    \multicolumn{1}{|l|}{c19\_text}                                 & 0.43          & 0.64          & {\ul 0.79}    & \textbf{0.94}   \\
    \multicolumn{1}{|l|}{miscov}                                    & {\ul 0.54}    & 0.45          & 0.47          & \textbf{0.57}   \\
    \multicolumn{1}{|l|}{c19\_title}                                & 0.49          & 0.75          & {\ul 0.77}    & \textbf{0.90}   \\ \hline
    \end{tabular}
    \end{table}

% \usepackage{multirow}
% \usepackage[normalem]{ulem}
\begin{table}[]
    \caption{Generalization improvement with KMP for model trained on `rumor'}
    \label{tab:rumor}
    \begin{tabular}{|lrrrr|}
    \hline
    \multicolumn{5}{|c|}{\textbf{Trained on `rumor'}}                                                                                   \\ \hline
    \multicolumn{1}{|c|}{\multirow{2}{*}{\textbf{Testing Dataset}}} & \multicolumn{4}{c|}{\textbf{Approach}}                          \\ \cline{2-5} 
    \multicolumn{1}{|c|}{}                                          & SB            & SE            & FT            & \textbf{FT+KMP} \\ \hline
    \multicolumn{1}{|l|}{cov\_fn}                                   & {\ul 0.76}    & 0.48          & 0.75          & \textbf{0.77}   \\
    \multicolumn{1}{|l|}{k\_short}                                  & 0.49          & {\ul 0.52}    & 0.52          & \textbf{0.54}   \\
    \multicolumn{1}{|l|}{coaid}                                     & 0.17          & 0.17          & {\ul 0.40}    & \textbf{0.76}      \\
    \multicolumn{1}{|l|}{cq}                                        & 0.47          & \textbf{0.59} & 0.42          & {\ul 0.58}      \\
    \multicolumn{1}{|l|}{k\_long}                                   & \textbf{0.59} & 0.52          & 0.43          & {\ul 0.53}      \\
    \multicolumn{1}{|l|}{rumor}                                     & 0.70          & 0.66          & {\ul 0.83}    & \textbf{0.86}   \\
    \multicolumn{1}{|l|}{c19\_text}                                 & 0.17          & 0.57          & {\ul 0.58}    & \textbf{0.73}   \\
    \multicolumn{1}{|l|}{miscov}                                    & 0.52          & 0.45          & {\ul 0.54}    & \textbf{0.54}      \\
    \multicolumn{1}{|l|}{c19\_title}                                & 0.74          & 0.57          & \textbf{0.81} & {\ul 0.76}      \\ \hline
    \end{tabular}
    \end{table}

\subsection{KMeans-Proxy Results}
We now show results from using KMeans-Proxy as a reject option 
in \cref{tab:coaid} and \cref{tab:rumor}. 
With rejection, we can increase generalization 
accuracy on unseen data. 
This is a faster approach than domain adaptation, since proxies are updated 
during training. 
Further, it is a plug-and-play solution, allowing for faster iteration on 
overall model design.

With KMeans-Proxy, we are able to improve generalization performance across the board.
Here, we compare models trained on `coaid' and `rumor' in 
\cref{tab:coaid} and \cref{tab:rumor}, respectively. 
Models are compared across static backbone (SB), static embedding 
(SE), fine-tuned (FT), and fine-tuned with KMeans-Proxy (FT+KMP).

For models trained on `coaid' (in \cref{tab:coaid}) and tested on all datasets, 
incorporating KMeans-Proxy improves generalization performance. 
In each case, FT+KMP is either the best or the runner-up model 
by at most 0.05 f1 f1 points. 
We see similar result for models trained on `rumor' in \cref{tab:rumor},
where KMeans-Proxy is either the best model or runner up for every
testing dataset. 

\PP{Choice of Proxy Factor}
Increasing the proxy factor leads to better generalization performance. 
\cref{tab:proxy} shows performance of a model trained on `c19\_text` 
that has poor generalization without KMeans-Proxy (see \cref{fig:ftconf}). 
As we increase the proxy factor, we gain better generalization 
across testing datasets. 
We test with different proxy factors and compare generalization performance 
of each model. 
We find that increasing the proxy factor leads to small, but measurable 
increase in accuracy. 
%
%In turn, coverage drops from 100% to on average 70% across datasets.

% \usepackage{multirow}
\begin{table}[]
    \caption{Impact of changing proxy factor: increasing the proxy factor increases accuracy, since more proxies allow tighter bounds on local coverage estimates.}
    \label{tab:proxy}
    \begin{tabular}{|l|rrrrrr|}
    \hline
    \multicolumn{1}{|c|}{\multirow{2}{*}{\textbf{Testing Dataset}}} & \multicolumn{6}{c|}{\textbf{Trained on c19\_text}}                                      \\ \cline{2-7} 
    \multicolumn{1}{|c|}{}                                          & \textbf{FT} & \textbf{k=1} & \textbf{k=2} & \textbf{k=3} & \textbf{k=5} & \textbf{k=10} \\ \hline
    cov\_fn                                                         & 0.44        & 0.50         & 0.54         & 0.58         & 0.58         & \textbf{0.58} \\
    k\_short                                                        & 0.59        & 0.70         & 0.73         & 0.73         & 0.73         & \textbf{0.75} \\
    coaid                                                           & 0.76        & 0.88         & 0.84         & 0.89         & 0.89         & \textbf{0.90} \\
    cq                                                              & 0.51        & 0.56         & 0.58         & 0.58         & 0.60         & \textbf{0.63} \\
    k\_long                                                         & 0.60        & 0.70         & 0.72         & 0.73         & 0.73         & \textbf{0.73} \\
    rumor                                                           & 0.47        & 0.45         & 0.51         & 0.58         & 0.58         & \textbf{0.67} \\
    c19\_text                                                       & 0.97        & 0.98         & 0.99         & 0.99         & 0.99         & \textbf{0.99} \\
    miscov                                                          & 0.48        & 0.45         & 0.44         & 0.53         & 0.58         & \textbf{0.58} \\
    c19\_title                                                      & 0.58        & 0.61         & 0.64         & 0.69         & 0.68         & \textbf{0.73} \\ \hline
    \end{tabular}
    \end{table}

\subsection{Discussion}
There are several observations we can make from our
generalization studies and KMeans-Proxy experiments.
\PP{Generalization}
For fake news detection, fine-tuned models must be
used carefully to take advntage of learned parameters.
As we showed in the confusion matrices, fine-tuning 
improves performance only on subsets of unseen data.
These subsets are regions of the data space where
the unseen data overlaps with training data.
On completely new regions of the data space, fine-tuned 
models make mistakes.
These mistakes are because of label overlap.

We must make a distinction between label and data
overlap.
Data overlap means a model has coverage on the unseen
data, and can make predictions with higher confidence.
Label coverage, as we showed in Fig, indicates where 
different labels occur close to each other in the 
embedding space.
Both can coincide: unseen data points can have both
data \textit{and} label overlap.
For these points, fine-tuned models that have
better captured a local region with training data 
are better poised to provide high-confidence labels.

\PP{KMeans-Proxy}
KMeans-Proxy allows us to identify these regions.
With KMeans-Proxy, we partition the data space into 
clusters representing model coverage and labels.
Our inclusion of the proxy factor $k$, where we create $k$
clusters for each class label, allows fine-grained partitioning
of the embedding space.
This means we can better capture local characteristics 
of the embedding space \cite{foundation2}.
In our experiments, we focus on 2 such characteristics:
(i) whether the label for an unseen point matches
the label for nearest proxy, and
(ii) whether this unseen point is within one 
standard-deviation radius of the proxy.
In our experiments, we show that using these provides
improvements in generalizing to unseen data points.

Clearly, there is significant progress to be made in
capturing local characteristics.
For example, when using an ensemble of fine-tuned models,
local smoothness \cite{niceness, foundation2} can be computed for each 
non-abstaining model to rank them on coverage.
There may also be advantages in using dynamic proxy allocation.
If prior class balance is known, then we could use a class-specific
proxy factor.

\section{Conclusion}
\label{sec:conclusion}
In this paper, we have presented generalizability
experiments and KMeans-Proxy.
We perform generalization studies across 9 fake
news datasets using several transformer-based
fake news detector models.
Our generalizability experiments show that fine-tuned
models generalize well to unseen data when 
there is overlap between unseen and training data.
On unseen data that does not overlap, fine-tuned 
models make mistakes due to poor coverage, label
flipping, and concept drift.

Using our observations and recent research into 
local embedding regions, we develop and present
KMeans-Proxy, a simple online KMeans clusterer
paired with a proxy factor.
With KMeans-Proxy, we partition the embedding space into
local regions and use local characteristics to create a
reject option for models.
We show that KMeans-Proxy improves generalization accuracy
for fine-tuned models across all 9 fake news datasets.
We welcome future research in this area to better
explore the generalizability and fine-tuning tradeoff.

%\newpage
% \input{ack}

\bibliographystyle{ACM-Reference-Format}
\bibliography{main}


\begin{thebibliography}{49}

%%% the Web address.

\ifx \showCODEN    \undefined \def \showCODEN     #1{\unskip}     \fi
\ifx \showDOI      \undefined \def \showDOI       #1{#1}\fi
\ifx \showISBNx    \undefined \def \showISBNx     #1{\unskip}     \fi
\ifx \showISBNxiii \undefined \def \showISBNxiii  #1{\unskip}     \fi
\ifx \showISSN     \undefined \def \showISSN      #1{\unskip}     \fi
\ifx \showLCCN     \undefined \def \showLCCN      #1{\unskip}     \fi
\ifx \shownote     \undefined \def \shownote      #1{#1}          \fi
\ifx \showarticletitle \undefined \def \showarticletitle #1{#1}   \fi
\ifx \showURL      \undefined \def \showURL       {\relax}        \fi
% invisible to TeX
\providecommand\bibfield[2]{#2}
\providecommand\bibinfo[2]{#2}
\providecommand\natexlab[1]{#1}
\providecommand\showeprint[2][]{arXiv:#2}

\bibitem[Agarwal(2020)]%
        {cov19fn}
\bibfield{author}{\bibinfo{person}{Isha Agarwal}.}
  \bibinfo{year}{2020}\natexlab{}.
\newblock \showarticletitle{{Covid 19 Fake News Dataset}}.
\newblock  (\bibinfo{date}{6} \bibinfo{year}{2020}).
\newblock
\urldef\tempurl%
\url{https://doi.org/10.6084/m9.figshare.12489293.v1}
\showDOI{\tempurl}


\bibitem[Bang et~al\mbox{.}(2021)]%
        {generalization2}
\bibfield{author}{\bibinfo{person}{Yejin Bang}, \bibinfo{person}{Etsuko Ishii},
  \bibinfo{person}{Samuel Cahyawijaya}, \bibinfo{person}{Ziwei Ji}, {and}
  \bibinfo{person}{Pascale Fung}.} \bibinfo{year}{2021}\natexlab{}.
\newblock \showarticletitle{Model generalization on covid-19 fake news
  detection}. In \bibinfo{booktitle}{\emph{International Workshop on Combating
  Hostile Posts in Regional Languages during Emergency Situation}}. Springer,
  \bibinfo{pages}{128--140}.
\newblock


\bibitem[Bommasani et~al\mbox{.}(2021)]%
        {foundation1}
\bibfield{author}{\bibinfo{person}{Rishi Bommasani}, \bibinfo{person}{Drew~A
  Hudson}, \bibinfo{person}{Ehsan Adeli}, \bibinfo{person}{Russ Altman},
  \bibinfo{person}{Simran Arora}, \bibinfo{person}{Sydney von Arx},
  \bibinfo{person}{Michael~S Bernstein}, \bibinfo{person}{Jeannette Bohg},
  \bibinfo{person}{Antoine Bosselut}, \bibinfo{person}{Emma Brunskill},
  {et~al\mbox{.}}} \bibinfo{year}{2021}\natexlab{}.
\newblock \showarticletitle{On the opportunities and risks of foundation
  models}.
\newblock \bibinfo{journal}{\emph{arXiv preprint arXiv:2108.07258}}
  (\bibinfo{year}{2021}).
\newblock


\bibitem[Brinkrolf and Hammer(2018)]%
        {reject2}
\bibfield{author}{\bibinfo{person}{Johannes Brinkrolf} {and}
  \bibinfo{person}{Barbara Hammer}.} \bibinfo{year}{2018}\natexlab{}.
\newblock \showarticletitle{Interpretable machine learning with reject option}.
\newblock \bibinfo{journal}{\emph{at-Automatisierungstechnik}}
  \bibinfo{volume}{66}, \bibinfo{number}{4} (\bibinfo{year}{2018}),
  \bibinfo{pages}{283--290}.
\newblock


\bibitem[Cardillo and L.~Warren(2016)]%
        {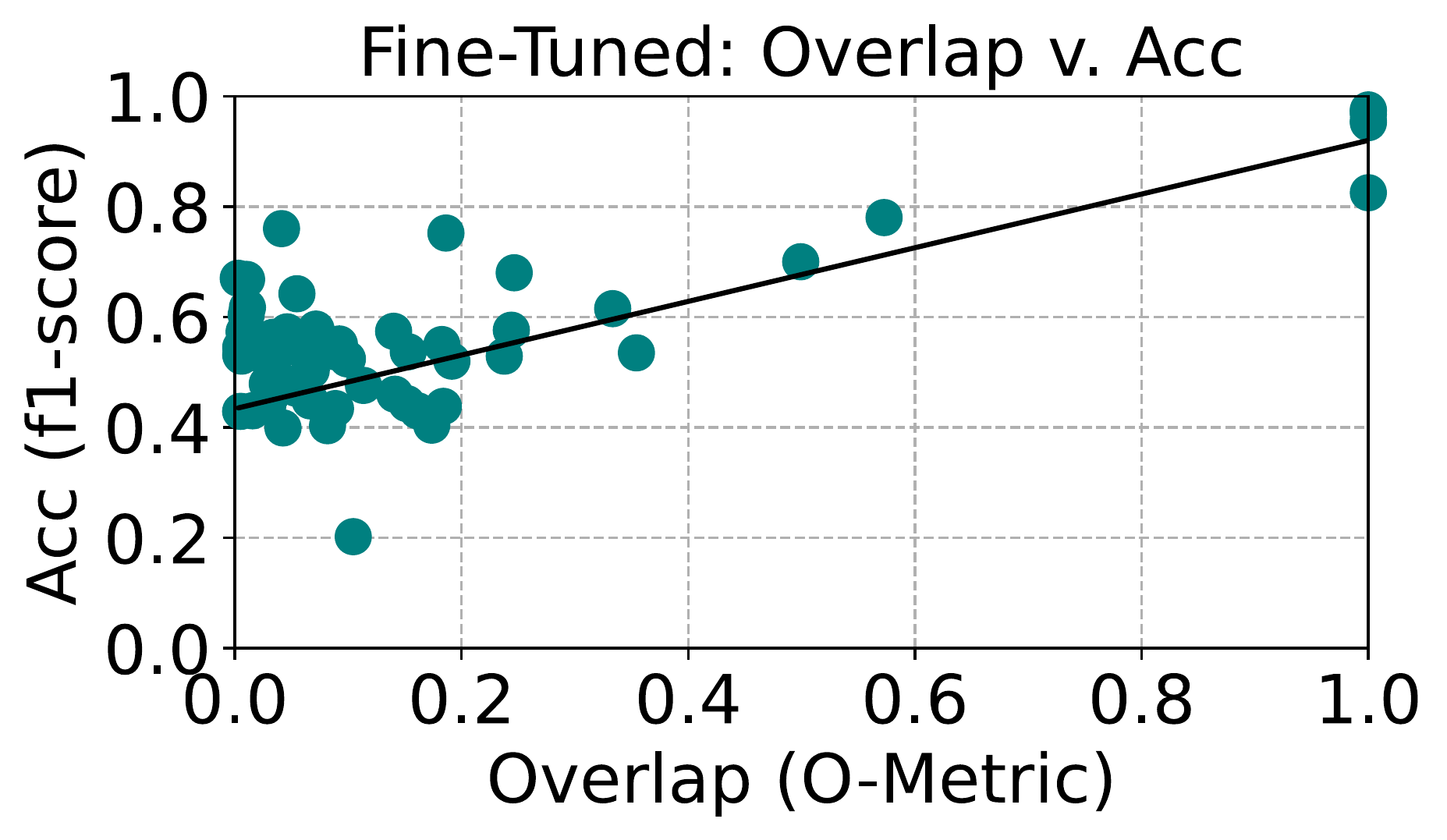}
\bibfield{author}{\bibinfo{person}{Marcel Cardillo} {and} \bibinfo{person}{Dan
  L.~Warren}.} \bibinfo{year}{2016}\natexlab{}.
\newblock \showarticletitle{Analysing patterns of spatial and niche overlap
  among species at multiple resolutions}.
\newblock \bibinfo{journal}{\emph{Global Ecology and Biogeography}}
  \bibinfo{volume}{25}, \bibinfo{number}{8} (\bibinfo{year}{2016}),
  \bibinfo{pages}{951--963}.
\newblock
\urldef\tempurl%
\url{https://doi.org/10.1111/geb.12455}
\showDOI{\tempurl}


\bibitem[Chen et~al\mbox{.}(2021)]%
        {fnft}
\bibfield{author}{\bibinfo{person}{Ben Chen}, \bibinfo{person}{Bin Chen},
  \bibinfo{person}{Dehong Gao}, \bibinfo{person}{Qijin Chen},
  \bibinfo{person}{Chengfu Huo}, \bibinfo{person}{Xiaonan Meng},
  \bibinfo{person}{Weijun Ren}, {and} \bibinfo{person}{Yang Zhou}.}
  \bibinfo{year}{2021}\natexlab{}.
\newblock \showarticletitle{Transformer-based language model fine-tuning
  methods for COVID-19 fake news detection}. In
  \bibinfo{booktitle}{\emph{International Workshop on Combating Hostile Posts
  in Regional Languages during Emergency Situation}}. Springer,
  \bibinfo{pages}{83--92}.
\newblock


\bibitem[Chen et~al\mbox{.}(2022)]%
        {foundation2}
\bibfield{author}{\bibinfo{person}{Mayee~F Chen}, \bibinfo{person}{Daniel~Y
  Fu}, \bibinfo{person}{Dyah Adila}, \bibinfo{person}{Michael Zhang},
  \bibinfo{person}{Frederic Sala}, \bibinfo{person}{Kayvon Fatahalian}, {and}
  \bibinfo{person}{Christopher R{\'e}}.} \bibinfo{year}{2022}\natexlab{}.
\newblock \showarticletitle{Shoring Up the Foundations: Fusing Model Embeddings
  and Weak Supervision}.
\newblock \bibinfo{journal}{\emph{arXiv preprint arXiv:2203.13270}}
  (\bibinfo{year}{2022}).
\newblock


\bibitem[Cheng et~al\mbox{.}(2021)]%
        {covrumor}
\bibfield{author}{\bibinfo{person}{Mingxi Cheng}, \bibinfo{person}{Songli
  Wang}, \bibinfo{person}{Xiaofeng Yan}, \bibinfo{person}{Tianqi Yang},
  \bibinfo{person}{Wenshuo Wang}, \bibinfo{person}{Zehao Huang},
  \bibinfo{person}{Xiongye Xiao}, \bibinfo{person}{Shahin Nazarian}, {and}
  \bibinfo{person}{Paul Bogdan}.} \bibinfo{year}{2021}\natexlab{}.
\newblock \showarticletitle{A COVID-19 Rumor Dataset}.
\newblock \bibinfo{journal}{\emph{Frontiers in Psychology}}
  \bibinfo{volume}{12} (\bibinfo{year}{2021}), \bibinfo{pages}{1566}.
\newblock


\bibitem[Cui and Lee(2020)]%
        {coaid}
\bibfield{author}{\bibinfo{person}{Limeng Cui} {and} \bibinfo{person}{Dongwon
  Lee}.} \bibinfo{year}{2020}\natexlab{}.
\newblock \showarticletitle{Coaid: Covid-19 healthcare misinformation dataset}.
\newblock \bibinfo{journal}{\emph{arXiv preprint arXiv:2006.00885}}
  (\bibinfo{year}{2020}).
\newblock


\bibitem[Das et~al\mbox{.}(2021)]%
        {covidfn}
\bibfield{author}{\bibinfo{person}{Sourya~Dipta Das}, \bibinfo{person}{Ayan
  Basak}, {and} \bibinfo{person}{Saikat Dutta}.}
  \bibinfo{year}{2021}\natexlab{}.
\newblock \showarticletitle{A heuristic-driven uncertainty based ensemble
  framework for fake news detection in tweets and news articles}.
\newblock \bibinfo{journal}{\emph{Neurocomputing}} (\bibinfo{year}{2021}).
\newblock


\bibitem[Devlin et~al\mbox{.}(2018)]%
        {bert}
\bibfield{author}{\bibinfo{person}{Jacob Devlin}, \bibinfo{person}{Ming-Wei
  Chang}, \bibinfo{person}{Kenton Lee}, {and} \bibinfo{person}{Kristina
  Toutanova}.} \bibinfo{year}{2018}\natexlab{}.
\newblock \showarticletitle{BERT: Pre-training of Deep Bidirectional
  Transformers for Language Understanding}.
\newblock \bibinfo{journal}{\emph{arXiv preprint arXiv:1810.04805}}
  (\bibinfo{year}{2018}).
\newblock


\bibitem[Enders et~al\mbox{.}(2020a)]%
        {misinfo}
\bibfield{author}{\bibinfo{person}{Adam~M Enders}, \bibinfo{person}{Joseph~E
  Uscinski}, \bibinfo{person}{Casey Klofstad}, {and} \bibinfo{person}{Justin
  Stoler}.} \bibinfo{year}{2020}\natexlab{a}.
\newblock \showarticletitle{The different forms of COVID-19 misinformation and
  their consequences}.
\newblock \bibinfo{journal}{\emph{The Harvard Kennedy School Misinformation
  Review}} (\bibinfo{year}{2020}).
\newblock


\bibitem[Enders et~al\mbox{.}(2020b)]%
        {infodemic}
\bibfield{author}{\bibinfo{person}{Adam~M Enders}, \bibinfo{person}{Joseph~E
  Uscinski}, \bibinfo{person}{Casey Klofstad}, {and} \bibinfo{person}{Justin
  Stoler}.} \bibinfo{year}{2020}\natexlab{b}.
\newblock \showarticletitle{The different forms of COVID-19 misinformation and
  their consequences}.
\newblock \bibinfo{journal}{\emph{The Harvard Kennedy School Misinformation
  Review}} (\bibinfo{year}{2020}).
\newblock


\bibitem[Gama et~al\mbox{.}(2014)]%
        {drift1}
\bibfield{author}{\bibinfo{person}{Jo{\~a}o Gama}, \bibinfo{person}{Indr{\.e}
  {\v{Z}}liobait{\.e}}, \bibinfo{person}{Albert Bifet}, \bibinfo{person}{Mykola
  Pechenizkiy}, {and} \bibinfo{person}{Abdelhamid Bouchachia}.}
  \bibinfo{year}{2014}\natexlab{}.
\newblock \showarticletitle{A survey on concept drift adaptation}.
\newblock \bibinfo{journal}{\emph{ACM computing surveys (CSUR)}}
  \bibinfo{volume}{46}, \bibinfo{number}{4} (\bibinfo{year}{2014}),
  \bibinfo{pages}{1--37}.
\newblock


\bibitem[Geifman and El-Yaniv(2019)]%
        {reject3}
\bibfield{author}{\bibinfo{person}{Yonatan Geifman} {and} \bibinfo{person}{Ran
  El-Yaniv}.} \bibinfo{year}{2019}\natexlab{}.
\newblock \showarticletitle{Selectivenet: A deep neural network with an
  integrated reject option}. In \bibinfo{booktitle}{\emph{International
  Conference on Machine Learning}}. PMLR, \bibinfo{pages}{2151--2159}.
\newblock


\bibitem[Hendrickx et~al\mbox{.}(2021)]%
        {reject1}
\bibfield{author}{\bibinfo{person}{Kilian Hendrickx}, \bibinfo{person}{Lorenzo
  Perini}, \bibinfo{person}{Dries Van~der Plas}, \bibinfo{person}{Wannes
  Meert}, {and} \bibinfo{person}{Jesse Davis}.}
  \bibinfo{year}{2021}\natexlab{}.
\newblock \showarticletitle{Machine Learning with a Reject Option: A survey}.
\newblock \bibinfo{journal}{\emph{arXiv preprint arXiv:2107.11277}}
  (\bibinfo{year}{2021}).
\newblock


\bibitem[Hossain et~al\mbox{.}(2020)]%
        {colies}
\bibfield{author}{\bibinfo{person}{Tamanna Hossain}, \bibinfo{person}{Robert~L
  Logan~IV}, \bibinfo{person}{Arjuna Ugarte}, \bibinfo{person}{Yoshitomo
  Matsubara}, \bibinfo{person}{Sean Young}, {and} \bibinfo{person}{Sameer
  Singh}.} \bibinfo{year}{2020}\natexlab{}.
\newblock \showarticletitle{COVIDLies: Detecting COVID-19 misinformation on
  social media}.
\newblock  (\bibinfo{year}{2020}).
\newblock


\bibitem[Hulburd(2020)]%
        {sefreezing2}
\bibfield{author}{\bibinfo{person}{Eric Hulburd}.}
  \bibinfo{year}{2020}\natexlab{}.
\newblock \showarticletitle{Exploring BERT Parameter Efficiency on the Stanford
  Question Answering Dataset v2.0}.
\newblock  (\bibinfo{year}{2020}).
\newblock
\urldef\tempurl%
\url{https://doi.org/10.48550/ARXIV.2002.10670}
\showDOI{\tempurl}


\bibitem[Jiao et~al\mbox{.}(2021)]%
        {sefreezing4}
\bibfield{author}{\bibinfo{person}{Xiaoqi Jiao}, \bibinfo{person}{Yichun Yin},
  \bibinfo{person}{Lifeng Shang}, \bibinfo{person}{Xin Jiang},
  \bibinfo{person}{Xiao Chen}, \bibinfo{person}{Linlin Li},
  \bibinfo{person}{Fang Wang}, {and} \bibinfo{person}{Qun Liu}.}
  \bibinfo{year}{2021}\natexlab{}.
\newblock \bibinfo{title}{LightMBERT: A Simple Yet Effective Method for
  Multilingual BERT Distillation}.
\newblock
\newblock
\urldef\tempurl%
\url{https://doi.org/10.48550/ARXIV.2103.06418}
\showDOI{\tempurl}


\bibitem[Kaliyar et~al\mbox{.}(2021)]%
        {mcnnet}
\bibfield{author}{\bibinfo{person}{Rohit~Kumar Kaliyar},
  \bibinfo{person}{Anurag Goswami}, {and} \bibinfo{person}{Pratik Narang}.}
  \bibinfo{year}{2021}\natexlab{}.
\newblock \showarticletitle{MCNNet: generalizing Fake News Detection with a
  Multichannel Convolutional Neural Network using a Novel COVID-19 Dataset}.
\newblock In \bibinfo{booktitle}{\emph{8th ACM IKDD CODS and 26th COMAD}}.
  \bibinfo{pages}{437--437}.
\newblock


\bibitem[Kou et~al\mbox{.}(2021)]%
        {cosocial}
\bibfield{author}{\bibinfo{person}{Ziyi Kou}, \bibinfo{person}{Lanyu Shang},
  \bibinfo{person}{Yang Zhang}, \bibinfo{person}{Christina Youn}, {and}
  \bibinfo{person}{Dong Wang}.} \bibinfo{year}{2021}\natexlab{}.
\newblock \showarticletitle{Fakesens: A social sensing approach to covid-19
  misinformation detection on social media}. In \bibinfo{booktitle}{\emph{2021
  17th International Conference on Distributed Computing in Sensor Systems
  (DCOSS)}}. IEEE, \bibinfo{pages}{140--147}.
\newblock


\bibitem[Kouw and Loog(2018)]%
        {drift2}
\bibfield{author}{\bibinfo{person}{Wouter~M Kouw} {and} \bibinfo{person}{Marco
  Loog}.} \bibinfo{year}{2018}\natexlab{}.
\newblock \showarticletitle{An introduction to domain adaptation and transfer
  learning}.
\newblock \bibinfo{journal}{\emph{arXiv preprint arXiv:1812.11806}}
  (\bibinfo{year}{2018}).
\newblock


\bibitem[Lan et~al\mbox{.}(2019)]%
        {albert}
\bibfield{author}{\bibinfo{person}{Zhenzhong Lan}, \bibinfo{person}{Mingda
  Chen}, \bibinfo{person}{Sebastian Goodman}, \bibinfo{person}{Kevin Gimpel},
  \bibinfo{person}{Piyush Sharma}, {and} \bibinfo{person}{Radu Soricut}.}
  \bibinfo{year}{2019}\natexlab{}.
\newblock \showarticletitle{Albert: A lite bert for self-supervised learning of
  language representations}.
\newblock \bibinfo{journal}{\emph{arXiv preprint arXiv:1909.11942}}
  (\bibinfo{year}{2019}).
\newblock


\bibitem[Lazer et~al\mbox{.}(2014)]%
        {gft}
\bibfield{author}{\bibinfo{person}{David Lazer}, \bibinfo{person}{Ryan
  Kennedy}, \bibinfo{person}{Gary King}, {and} \bibinfo{person}{Alessandro
  Vespignani}.} \bibinfo{year}{2014}\natexlab{}.
\newblock \showarticletitle{Google Flu Trends still appears sick: An evaluation
  of the 2013-2014 flu season}.
\newblock \bibinfo{journal}{\emph{Available at SSRN 2408560}}
  (\bibinfo{year}{2014}).
\newblock


\bibitem[Lee et~al\mbox{.}(2019)]%
        {sefreezing1}
\bibfield{author}{\bibinfo{person}{Jaejun Lee}, \bibinfo{person}{Raphael Tang},
  {and} \bibinfo{person}{Jimmy Lin}.} \bibinfo{year}{2019}\natexlab{}.
\newblock \showarticletitle{What Would Elsa Do? Freezing Layers During
  Transformer Fine-Tuning}.
\newblock  (\bibinfo{year}{2019}).
\newblock
\urldef\tempurl%
\url{https://doi.org/10.48550/ARXIV.1911.03090}
\showDOI{\tempurl}


\bibitem[Li et~al\mbox{.}(2021)]%
        {mdaws}
\bibfield{author}{\bibinfo{person}{Yichuan Li}, \bibinfo{person}{Kyumin Lee},
  \bibinfo{person}{Nima Kordzadeh}, \bibinfo{person}{Brenton Faber},
  \bibinfo{person}{Cameron Fiddes}, \bibinfo{person}{Elaine Chen}, {and}
  \bibinfo{person}{Kai Shu}.} \bibinfo{year}{2021}\natexlab{}.
\newblock \showarticletitle{Multi-Source Domain Adaptation with Weak
  Supervision for Early Fake News Detection}. In \bibinfo{booktitle}{\emph{2021
  IEEE International Conference on Big Data (Big Data)}}. IEEE,
  \bibinfo{pages}{668--676}.
\newblock


\bibitem[Memon and Carley(2020)]%
        {miscov}
\bibfield{author}{\bibinfo{person}{Shahan~Ali Memon} {and}
  \bibinfo{person}{Kathleen~M Carley}.} \bibinfo{year}{2020}\natexlab{}.
\newblock \showarticletitle{Characterizing covid-19 misinformation communities
  using a novel twitter dataset}.
\newblock \bibinfo{journal}{\emph{arXiv preprint arXiv:2008.00791}}
  (\bibinfo{year}{2020}).
\newblock


\bibitem[Merchant et~al\mbox{.}(2020)]%
        {seimportant}
\bibfield{author}{\bibinfo{person}{Amil Merchant}, \bibinfo{person}{Elahe
  Rahimtoroghi}, \bibinfo{person}{Ellie Pavlick}, {and} \bibinfo{person}{Ian
  Tenney}.} \bibinfo{year}{2020}\natexlab{}.
\newblock \bibinfo{title}{What Happens To BERT Embeddings During Fine-tuning?}
\newblock
\newblock
\urldef\tempurl%
\url{https://doi.org/10.48550/ARXIV.2004.14448}
\showDOI{\tempurl}


\bibitem[Moon et~al\mbox{.}(2019)]%
        {sefreezing3}
\bibfield{author}{\bibinfo{person}{Taesun Moon}, \bibinfo{person}{Parul
  Awasthy}, \bibinfo{person}{Jian Ni}, {and} \bibinfo{person}{Radu Florian}.}
  \bibinfo{year}{2019}\natexlab{}.
\newblock \showarticletitle{Towards Lingua Franca Named Entity Recognition with
  BERT}.
\newblock  (\bibinfo{year}{2019}).
\newblock
\urldef\tempurl%
\url{https://doi.org/10.48550/ARXIV.1912.01389}
\showDOI{\tempurl}


\bibitem[Movshovitz-Attias et~al\mbox{.}(2017)]%
        {proxy1}
\bibfield{author}{\bibinfo{person}{Yair Movshovitz-Attias},
  \bibinfo{person}{Alexander Toshev}, \bibinfo{person}{Thomas~K Leung},
  \bibinfo{person}{Sergey Ioffe}, {and} \bibinfo{person}{Saurabh Singh}.}
  \bibinfo{year}{2017}\natexlab{}.
\newblock \showarticletitle{No fuss distance metric learning using proxies}. In
  \bibinfo{booktitle}{\emph{Proceedings of the IEEE International Conference on
  Computer Vision}}. \bibinfo{pages}{360--368}.
\newblock


\bibitem[M{\"u}ller et~al\mbox{.}(2020)]%
        {covbert}
\bibfield{author}{\bibinfo{person}{Martin M{\"u}ller}, \bibinfo{person}{Marcel
  Salath{\'e}}, {and} \bibinfo{person}{Per~E Kummervold}.}
  \bibinfo{year}{2020}\natexlab{}.
\newblock \showarticletitle{Covid-twitter-bert: A natural language processing
  model to analyse covid-19 content on twitter}.
\newblock \bibinfo{journal}{\emph{arXiv preprint arXiv:2005.07503}}
  (\bibinfo{year}{2020}).
\newblock


\bibitem[Mutlu et~al\mbox{.}(2020)]%
        {covidcq}
\bibfield{author}{\bibinfo{person}{Ece~C Mutlu}, \bibinfo{person}{Toktam
  Oghaz}, \bibinfo{person}{Jasser Jasser}, \bibinfo{person}{Ege Tutunculer},
  \bibinfo{person}{Amirarsalan Rajabi}, \bibinfo{person}{Aida Tayebi},
  \bibinfo{person}{Ozlem Ozmen}, {and} \bibinfo{person}{Ivan Garibay}.}
  \bibinfo{year}{2020}\natexlab{}.
\newblock \showarticletitle{A Stance Data Set on Polarized Conversations on
  Twitter about the Efficacy of Hydroxychloroquine as a Treatment for
  COVID-19}.
\newblock \bibinfo{journal}{\emph{Data in Brief}} (\bibinfo{year}{2020}),
  \bibinfo{pages}{106401}.
\newblock


\bibitem[Orr et~al\mbox{.}(2021)]%
        {foundation3}
\bibfield{author}{\bibinfo{person}{Laurel Orr}, \bibinfo{person}{Karan Goel},
  {and} \bibinfo{person}{Christopher R{\'e}}.} \bibinfo{year}{2021}\natexlab{}.
\newblock \showarticletitle{Data management opportunities for foundation
  models}. In \bibinfo{booktitle}{\emph{12th Annual Conference on Innovative
  Data Systems Research}}.
\newblock


\bibitem[Patel(2021)]%
        {kagglefn}
\bibfield{author}{\bibinfo{person}{Sameer Patel}.}
  \bibinfo{year}{2021}\natexlab{}.
\newblock \showarticletitle{{Covid-19 Fake News Dataset (Kaggle)}}.
\newblock  (\bibinfo{year}{2021}).
\newblock
\urldef\tempurl%
\url{https://www.kaggle.com/code/therealsampat/fake-news-detection/}
\showURL{%
\tempurl}


\bibitem[Pu et~al\mbox{.}(2020)]%
        {ebka}
\bibfield{author}{\bibinfo{person}{Calton Pu}, \bibinfo{person}{Abhijit
  Suprem}, \bibinfo{person}{Rodrigo~Alves Lima}, \bibinfo{person}{Aibek
  Musaev}, \bibinfo{person}{De Wang}, \bibinfo{person}{Danesh Irani},
  \bibinfo{person}{Steve Webb}, {and} \bibinfo{person}{Joao~Eduardo Ferreira}.}
  \bibinfo{year}{2020}\natexlab{}.
\newblock \showarticletitle{Beyond artificial reality: Finding and monitoring
  live events from social sensors}.
\newblock \bibinfo{journal}{\emph{ACM Transactions on Internet Technology
  (TOIT)}} \bibinfo{volume}{20}, \bibinfo{number}{1} (\bibinfo{year}{2020}),
  \bibinfo{pages}{1--21}.
\newblock


\bibitem[Quinn et~al\mbox{.}(2021)]%
        {infolife}
\bibfield{author}{\bibinfo{person}{Emma~K Quinn}, \bibinfo{person}{Sajjad~S
  Fazel}, {and} \bibinfo{person}{Cheryl~E Peters}.}
  \bibinfo{year}{2021}\natexlab{}.
\newblock \showarticletitle{The Instagram infodemic: cobranding of conspiracy
  theories, coronavirus disease 2019 and authority-questioning beliefs}.
\newblock \bibinfo{journal}{\emph{Cyberpsychology, Behavior, and Social
  Networking}} \bibinfo{volume}{24}, \bibinfo{number}{8}
  (\bibinfo{year}{2021}), \bibinfo{pages}{573--577}.
\newblock


\bibitem[Ratner et~al\mbox{.}(2017)]%
        {snorkel}
\bibfield{author}{\bibinfo{person}{Alexander Ratner},
  \bibinfo{person}{Stephen~H Bach}, \bibinfo{person}{Henry Ehrenberg},
  \bibinfo{person}{Jason Fries}, \bibinfo{person}{Sen Wu}, {and}
  \bibinfo{person}{Christopher R{\'e}}.} \bibinfo{year}{2017}\natexlab{}.
\newblock \showarticletitle{Snorkel: Rapid training data creation with weak
  supervision}. In \bibinfo{booktitle}{\emph{Proceedings of the VLDB Endowment.
  International Conference on Very Large Data Bases}},
  Vol.~\bibinfo{volume}{11}. NIH Public Access, \bibinfo{pages}{269}.
\newblock


\bibitem[R{\"u}hling~Cachay et~al\mbox{.}(2021)]%
        {eews}
\bibfield{author}{\bibinfo{person}{Salva R{\"u}hling~Cachay},
  \bibinfo{person}{Benedikt Boecking}, {and} \bibinfo{person}{Artur
  Dubrawski}.} \bibinfo{year}{2021}\natexlab{}.
\newblock \showarticletitle{End-to-End Weak Supervision}.
\newblock \bibinfo{journal}{\emph{Advances in Neural Information Processing
  Systems}}  \bibinfo{volume}{34} (\bibinfo{year}{2021}).
\newblock


\bibitem[Serrano et~al\mbox{.}(2020)]%
        {fndetect}
\bibfield{author}{\bibinfo{person}{Juan Carlos~Medina Serrano},
  \bibinfo{person}{Orestis Papakyriakopoulos}, {and} \bibinfo{person}{Simon
  Hegelich}.} \bibinfo{year}{2020}\natexlab{}.
\newblock \showarticletitle{NLP-based feature extraction for the detection of
  COVID-19 misinformation videos on YouTube}. In
  \bibinfo{booktitle}{\emph{Proceedings of the 1st Workshop on NLP for COVID-19
  at ACL 2020}}.
\newblock


\bibitem[So et~al\mbox{.}(2022)]%
        {kmeans}
\bibfield{author}{\bibinfo{person}{Geelon So}, \bibinfo{person}{Gaurav
  Mahajan}, {and} \bibinfo{person}{Sanjoy Dasgupta}.}
  \bibinfo{year}{2022}\natexlab{}.
\newblock \showarticletitle{Convergence of online k-means}. In
  \bibinfo{booktitle}{\emph{International Conference on Artificial Intelligence
  and Statistics}}. PMLR, \bibinfo{pages}{8534--8569}.
\newblock


\bibitem[Suprem et~al\mbox{.}(2020)]%
        {odin}
\bibfield{author}{\bibinfo{person}{Abhijit Suprem}, \bibinfo{person}{Joy
  Arulraj}, \bibinfo{person}{Calton Pu}, {and} \bibinfo{person}{Joao
  Ferreira}.} \bibinfo{year}{2020}\natexlab{}.
\newblock \showarticletitle{Odin: automated drift detection and recovery in
  video analytics}.
\newblock \bibinfo{journal}{\emph{arXiv preprint arXiv:2009.05440}}
  (\bibinfo{year}{2020}).
\newblock


\bibitem[Suprem et~al\mbox{.}(2019)]%
        {litmus1}
\bibfield{author}{\bibinfo{person}{Abhijit Suprem}, \bibinfo{person}{Aibek
  Musaev}, {and} \bibinfo{person}{Calton Pu}.} \bibinfo{year}{2019}\natexlab{}.
\newblock \showarticletitle{Concept drift adaptive physical event detection for
  social media streams}. In \bibinfo{booktitle}{\emph{World Congress on
  Services}}. Springer, \bibinfo{pages}{92--105}.
\newblock


\bibitem[Urner and Ben-David(2013)]%
        {niceness}
\bibfield{author}{\bibinfo{person}{Ruth Urner} {and} \bibinfo{person}{Shai
  Ben-David}.} \bibinfo{year}{2013}\natexlab{}.
\newblock \showarticletitle{Probabilistic lipschitzness a niceness assumption
  for deterministic labels}. In \bibinfo{booktitle}{\emph{Learning Faster from
  Easy Data-Workshop NIPS}}, Vol.~\bibinfo{volume}{2}. \bibinfo{pages}{1}.
\newblock


\bibitem[Wahle et~al\mbox{.}(2022)]%
        {generalization}
\bibfield{author}{\bibinfo{person}{Jan~Philip Wahle}, \bibinfo{person}{Nischal
  Ashok}, \bibinfo{person}{Terry Ruas}, \bibinfo{person}{Norman Meuschke},
  \bibinfo{person}{Tirthankar Ghosal}, {and} \bibinfo{person}{Bela Gipp}.}
  \bibinfo{year}{2022}\natexlab{}.
\newblock \showarticletitle{Testing the generalization of neural language
  models for COVID-19 misinformation detection}. In
  \bibinfo{booktitle}{\emph{International Conference on Information}}.
  Springer, \bibinfo{pages}{381--392}.
\newblock


\bibitem[Walambe et~al\mbox{.}(2022)]%
        {explainmisinfo}
\bibfield{author}{\bibinfo{person}{Rahee Walambe}, \bibinfo{person}{Ananya
  Srivastava}, \bibinfo{person}{Bhargav Yagnik}, \bibinfo{person}{Mohammed
  Hasan}, \bibinfo{person}{Zainuddin Saiyed}, \bibinfo{person}{Gargi Joshi},
  {and} \bibinfo{person}{Ketan Kotecha}.} \bibinfo{year}{2022}\natexlab{}.
\newblock \showarticletitle{Explainable Misinformation Detection Across
  Multiple Social Media Platforms}.
\newblock \bibinfo{journal}{\emph{arXiv preprint arXiv:2203.11724}}
  (\bibinfo{year}{2022}).
\newblock


\bibitem[Weinzierl et~al\mbox{.}(2021)]%
        {misinfo1}
\bibfield{author}{\bibinfo{person}{Maxwell Weinzierl}, \bibinfo{person}{Suellen
  Hopfer}, {and} \bibinfo{person}{Sanda~M Harabagiu}.}
  \bibinfo{year}{2021}\natexlab{}.
\newblock \showarticletitle{Misinformation adoption or rejection in the era of
  covid-19}. In \bibinfo{booktitle}{\emph{Proceedings of the International AAAI
  Conference on Web and Social Media (ICWSM), AAAI Press}}.
\newblock


\bibitem[Wern~Teh et~al\mbox{.}(2020)]%
        {proxy2}
\bibfield{author}{\bibinfo{person}{Eu Wern~Teh}, \bibinfo{person}{Terrance
  DeVries}, {and} \bibinfo{person}{Graham~W Taylor}.}
  \bibinfo{year}{2020}\natexlab{}.
\newblock \showarticletitle{ProxyNCA++: Revisiting and Revitalizing Proxy
  Neighborhood Component Analysis}.
\newblock \bibinfo{journal}{\emph{arXiv e-prints}} (\bibinfo{year}{2020}),
  \bibinfo{pages}{arXiv--2004}.
\newblock


\bibitem[Yang et~al\mbox{.}(2015)]%
        {argo}
\bibfield{author}{\bibinfo{person}{Shihao Yang}, \bibinfo{person}{Mauricio
  Santillana}, {and} \bibinfo{person}{Samuel~C Kou}.}
  \bibinfo{year}{2015}\natexlab{}.
\newblock \showarticletitle{Accurate estimation of influenza epidemics using
  Google search data via ARGO}.
\newblock \bibinfo{journal}{\emph{Proceedings of the National Academy of
  Sciences}} \bibinfo{volume}{112}, \bibinfo{number}{47}
  (\bibinfo{year}{2015}), \bibinfo{pages}{14473--14478}.
\newblock


\bibitem[{\v{Z}}liobait{\.e}(2010)]%
        {drift3}
\bibfield{author}{\bibinfo{person}{Indr{\.e} {\v{Z}}liobait{\.e}}.}
  \bibinfo{year}{2010}\natexlab{}.
\newblock \showarticletitle{Learning under concept drift: an overview}.
\newblock \bibinfo{journal}{\emph{arXiv preprint arXiv:1010.4784}}
  (\bibinfo{year}{2010}).
\newblock


\end{thebibliography}

\end{document}